\documentclass[10pt,letterpaper]{article}
\providecommand{\WithCVPRCaseVersion}{0}

\providecommand{\WithCVPRCaseVersion}{0}
\newif\ifWithCVPRCases
\ifnum\WithCVPRCaseVersion=1\relax
  \WithCVPRCasestrue
\else
  \WithCVPRCasesfalse
\fi
\ifdefined\pdfminorversion\pdfminorversion=7\fi
\usepackage[T1]{fontenc}
\usepackage[pagenumbers]{westlakeagi}
\usepackage{tcolorbox}
\usepackage[pagebackref,breaklinks,colorlinks,allcolors=wagiblue]{hyperref}
\usepackage{colortbl}
\usepackage{array}
\usepackage{placeins}
\usepackage{float}
\usepackage{flafter}

\definecolor{oursred}{RGB}{235,244,250}
\providecommand{\Description}[1]{}
\newif\ifshowchanges
\showchangesfalse
\definecolor{revisionred}{RGB}{190,0,0}
\newcommand{\revisioncolor}{%
  \ifshowchanges\color{revisionred}\hypersetup{allcolors=revisionred}\fi}
\newenvironment{revision}{\revisioncolor}{\par}
\DeclareRobustCommand{\rev}[1]{\ifshowchanges{\revisioncolor#1}\else#1\fi}
\newcommand{\revisionfloat}{%
  \revisioncolor
  \ifshowchanges\captionsetup{font={small,color=revisionred}}\fi}
\expandafter\def\csname revisionref@asim25met3r\endcsname{}
\expandafter\def\csname revisionref@leroy2024grounding\endcsname{}
\expandafter\def\csname revisionref@wei2025freeflux\endcsname{}
\expandafter\def\csname revisionref@chen2026contextflow\endcsname{}
\expandafter\def\csname revisionref@wang2026free\endcsname{}
\expandafter\def\csname revisionref@zhao2026dyweight\endcsname{}
\expandafter\def\csname revisionref@kerbl20233d\endcsname{}
\expandafter\def\csname revisionref@liu2024stylegaussian\endcsname{}
\expandafter\def\csname revisionref@wang2004ssim\endcsname{}
\newcommand{\revisionbibitem}[2][]{%
  \ifshowchanges
    \ifcsname revisionref@#2\endcsname
      \color{revisionred}\hypersetup{allcolors=revisionred}%
    \else
      \color{black}\hypersetup{allcolors=wagiblue}%
    \fi
  \fi
  \originalbibitem[#1]{#2}}
\newcommand{\papertitle}{\rev{VidEdit3D: Flexible 3D Gaussian Editing with Video Diffusion}}
\title{\papertitle}
\author{Yanming Yang, Chenxi Song, Tong Zhao, Juntao Xu, Wenhao Rao, Yiwen Chen, Yiwei Wang, Chi Zhang}
\date{}
\wagilogo{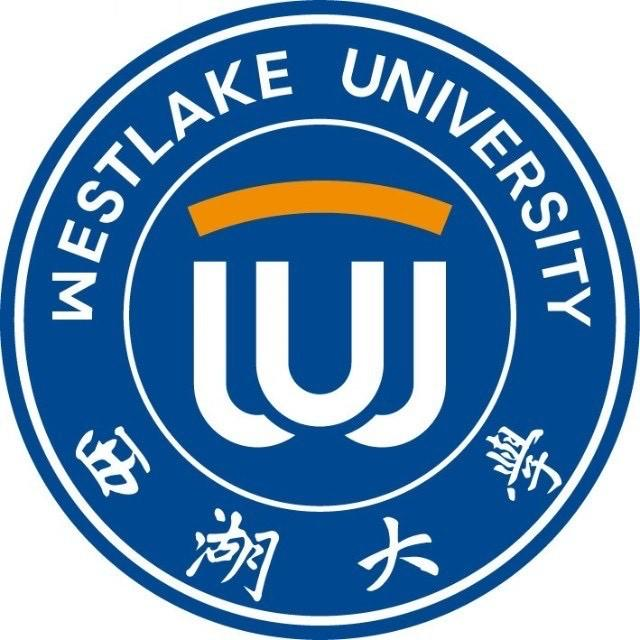}
\hypersetup{pdftitle={VidEdit3D: Flexible 3D Gaussian Editing with Video Diffusion},
  pdfauthor={Yanming Yang, Chenxi Song, Tong Zhao, Juntao Xu, Wenhao Rao, Yiwen Chen, Yiwei Wang, Chi Zhang},
  pdfsubject={Preprint}}
\begin{document}
\vspace*{-0.42in}
\wagibanner
\vskip 0.06in
\begin{tcolorbox}[colback=wagiblue!4,colframe=wagiblue!40,
  boxrule=0.6pt,arc=10pt,left=22pt,right=22pt,top=8pt,bottom=3pt]
\begingroup\centering
{\resizebox{\linewidth}{!}{\LARGE\bfseries \papertitle}\par}
\vskip 7pt
{\large\revisioncolor Yanming Yang\textsuperscript{1} \quad Chenxi Song\textsuperscript{1} \quad Tong Zhao\textsuperscript{1} \quad Juntao Xu\textsuperscript{1}\\
Wenhao Rao\textsuperscript{2} \quad Yiwen Chen\textsuperscript{3} \quad Yiwei Wang\textsuperscript{4} \quad Chi Zhang\textsuperscript{1$\dagger$}\par}
\vskip 3pt
{\revisioncolor\textsuperscript{1}Westlake University \quad \textsuperscript{2}University of Leeds\\
\textsuperscript{3}Nanyang Technological University \quad \textsuperscript{4}University of California, Merced\par}
\vskip 2pt
{\small\revisioncolor\textsuperscript{$\dagger$}Corresponding author.\par}
\vskip 3pt
{\small\rev{\textbf{Project Page:} \url{https://2hitee.github.io/VidEdit3D-page/}}\par}
\endgroup
\vskip 3pt
\renewenvironment{abstract}{\centerline{\large\bfseries Abstract}\vskip 4pt\fontsize{9.5}{11}\selectfont\itshape\noindent\ignorespaces}{\par}
\begin{abstract}
   Recent advancements in text-guided 3D editing, particularly with 3D Gaussian Splatting (3DGS), have shown promise but still lag behind the flexibility of 2D editing tools. Existing methods face a critical trade-off \rev{between preserving} multi-view consistency \rev{and allowing} structural modifications. \rev{Image-based editors can also struggle to apply the intended edit to objects viewed from unusual angles.} To address these limitations, we introduce \rev{VidEdit3D}, a \rev{framework} that reformulates 3D scene editing as a trajectory-based video editing task. By drawing a parallel between camera movement in static 3D scenes and temporal dynamics in video, we leverage the robust spatiotemporal priors of pre-trained Video Diffusion Models (VDMs) to \rev{encourage} semantic and structural coherence across views. Specifically, we propose Trajectory-Guided Rendering (TGR) to \rev{select informative keyframes and construct} smooth \rev{video} clips aligned with VDM input requirements. To propagate edits from an anchor view to these clips, we develop Flow-Aligned Video Editing (FAVE), a training-free framework incorporating \rev{an} Editing Consistency Guidance (ECG) mechanism. ECG \rev{contrasts source and edited image conditions at a shared latent state}, \rev{using scheduled source-image amplification to strengthen} adherence to the edited reference while maintaining original scene context. Finally, we introduce Region-Selective 3DGS Refinement (RS-GS) to strictly confine geometric and appearance updates to target primitives, \rev{reducing} background attribute leakage. \rev{Experiments demonstrate stronger instruction alignment and better multi-view consistency than the baselines.}
\end{abstract}

\end{tcolorbox}
\vskip 0.02in
\begingroup
\setlength{\intextsep}{2pt}
\begin{figure}[H]
\revisionfloat
\captionsetup{skip=4pt}
\centering
\includegraphics[width=\linewidth]{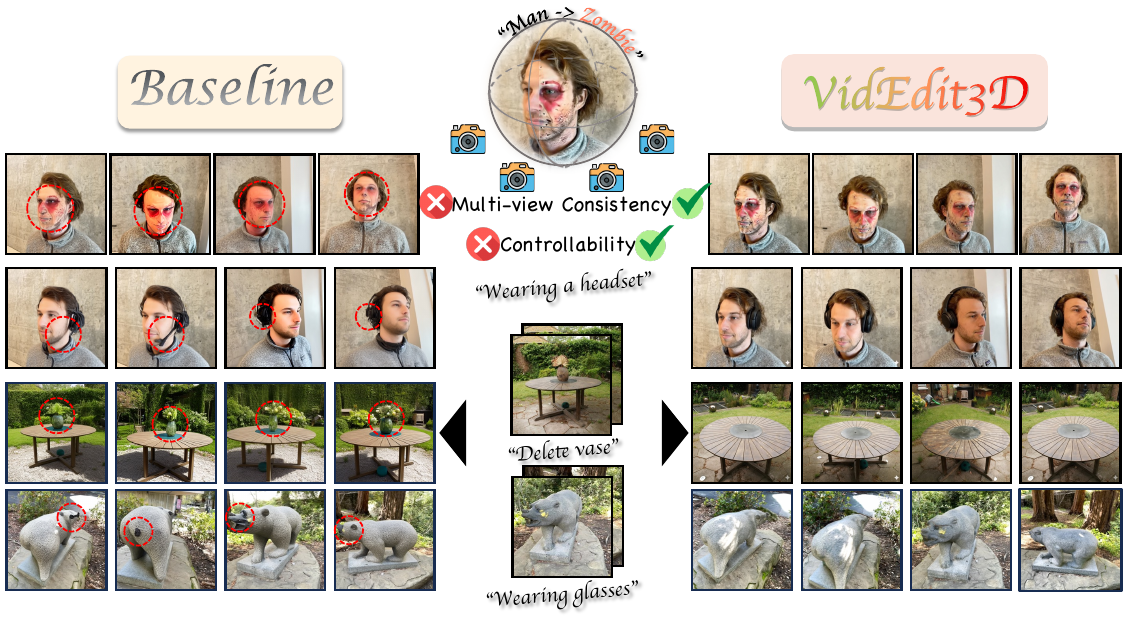}
\caption{Flexible and consistent 3D editing with VidEdit3D. The baseline (left) shows inconsistent appearance, incomplete headset insertion and vase removal, and misplaced glasses. VidEdit3D (right) propagates the anchor edit across views, producing coherent appearance changes and localized object insertion and removal while preserving nearby content.}
\label{fig:teaser}
\end{figure}
\endgroup

\section{Introduction}
\label{sec:intro}

Recent advances in text-guided image editing have demonstrated remarkable success in enabling users to modify images according to natural language instructions~\cite{brooks2023instructpix2pix,hertz2022prompt,tumanyan2023plug}.
By leveraging large-scale diffusion or transformer-based models, these systems can generate semantically aligned and visually coherent edits, such as replacing objects, altering styles, or modifying attributes, while preserving content fidelity.
This paradigm has significantly democratized content creation, making intuitive visual editing accessible to non-experts.
Building upon this progress, researchers have recently extended text-driven editing to the 3D domain, particularly in neural implicit representations such as Neural Radiance Fields (NeRFs)~\cite{mildenhall2021nerf} and explicit point-based representations such as 3D Gaussian Splatting (3DGS)~\cite{kerbl20233d}.
Particularly, compared with NeRF, 3DGS offers greater flexibility and efficiency as an explicit 3D representation, enabling high-quality real-time rendering and fine-grained local editing. These advantages make 3DGS editing an attractive paradigm for controllable 3D content manipulation, which is the focus of our paper.

However, despite the potential of 3DGS, the capabilities of existing 3D editing frameworks still lag significantly behind their 2D counterparts. \rev{Many} 3D editing methods \rev{focus on} operations \rev{such} as global style transfer, texture modification, or \rev{object} removal. This stands in stark contrast to state-of-the-art 2D editing tools, which support highly imaginative and complex transformations with ease. A primary bottleneck restricting 3D editing flexibility is the requirement for multi-view consistency; unlike 2D editing, modifying a 3D scene requires simultaneous adjustments across multiple viewpoints while ensuring the underlying geometry remains coherent. Early attempts that naively applied 2D editors independently across different views inherently lacked cross-view coherence~\cite{haque2023instruct, chen2024gaussianeditor, wang2024gaussianeditor}, leading to severe artifacts such as ghosting and flickering.
To mitigate these inconsistencies, subsequent works~\cite{wu2024gaussctrl, wang2024view, chen2024dge} introduced \rev{geometric guidance through} depth \rev{or correspondence} with the original images. While these constraints stabilize the optimization process, \rev{strong alignment to the source can limit} editing flexibility. This limitation becomes particularly evident in instructions that require structural changes. For instance, a simple instruction like ``make the person wear a hat'' becomes difficult to execute \rev{when} the required geometric addition conflicts with \rev{alignment} constraints derived from the original shape. 

\rev{Image-based editors can also struggle to apply the intended edit to objects viewed from unusual angles, especially when the relevant parts are partly or fully hidden.} \rev{For example, an instruction such as ``put glasses on a bear'' is readily interpreted from a frontal view but becomes ambiguous from a rear view, where the face is occluded. 3D editing therefore needs to propagate a clearly specified edit across views while accounting for which object parts are visible.}

These limitations motivate the design of a flexible 3D editing framework capable of performing precise semantic manipulations without the excessive rigidity of geometric locks. \rev{To achieve this, we introduce VidEdit3D (Fig.~\ref{fig:teaser}), which takes an edit created in a single view with a 2D editor, propagates it across viewpoints, and reconstructs the edited 3D scene.} To robustly facilitate this propagation, we identify a powerful parallel between temporal consistency in videos and multi-view consistency in the rendered frames of a 3D scene. A camera trajectory moving around a static 3D object yields a sequence of frames that is structurally analogous to a video clip. This insight inspires us to leverage the robust spatiotemporal priors of pre-trained Video Diffusion Models (VDMs)~\cite{wan2025wan, kong2024hunyuanvideo, blattmann2023stable, hong2022cogvideo, yang2024cogvideox}. \rev{VDMs} are \rev{designed} to maintain object identity and semantic coherence across time, \rev{providing a useful prior} for \rev{relating} 3D viewpoints. \rev{ViP3DE~\cite{chen2026fast} also propagates a single edited view into 3D using motion-preserved noise blending and geometry-aware denoising.} \rev{VidEdit3D strengthens reference transfer through an explicit contrast between source and edited image conditions, while spatial source regularization preserves the surrounding scene.} \rev{Our inversion-free formulation allows the edited region to change without enforcing source-geometry feature correspondences during denoising.} \rev{In our comparisons, VidEdit3D achieves stronger instruction alignment and better multi-view consistency than ViP3DE, including when both use the same video backbone.}

To transform sparse 3D renderings into a form compatible with video diffusion models, we introduce a Trajectory-Guided Rendering (TGR) strategy. \rev{Starting from a user-selected anchor, TGR selects views of the source subject carrying the edit and connects them into a smooth camera trajectory. We propagate along two branches from the anchor to shorten the sequence of transfers to distant views and reduce error accumulation. Overlapping clips maintain continuity within each branch while matching video diffusion priors.}

To propagate edits from a key view to the rendered video clips, we develop Flow-Aligned Video Editing (FAVE), a training-free framework inspired by recent inversion-free editing methods~\cite{kim2025flowalign, kulikov2025flowedit}. Unlike prior methods that are primarily guided by text, our goal is to leverage the strong generative and editing capabilities of 2D image editing models to transfer a user-specified edit from one anchor view to the full 3D scene.

In standard \rev{image-to-video (I2V)} generation, classifier-free guidance (CFG)~\cite{ho2022classifier} evaluates the denoiser twice under different text conditions, while both evaluations typically share the same first-frame image as the temporal anchor. This design is effective for video synthesis, since a fixed first-frame condition helps preserve object identity and temporal coherence over time. In contrast, FAVE, as an inversion-free editing method, conceptually requires two distinct conditioning roles: one should follow the edited reference frame and target prompt, while the other should preserve the original scene structure under the source frame and source prompt. However, directly reusing the standard I2V conditioning scheme in our setting leads both evaluations to remain anchored to the same edited first frame, rather than forming the desired source-versus-target contrast. In I2V models, this mismatch is exacerbated by the strong first-frame anchoring effect: \rev{later frames remain strongly influenced by the first-frame image condition}, \rev{which can make} the two evaluations \rev{inherit} similar image-conditioned components \rev{and weaken} the intended guidance contrast\rev{.} To address this issue, we introduce Editing Consistency Guidance (ECG), a training-free inference-time guidance scheme that restores a meaningful source-versus-edit contrast in the guidance computation. \rev{Specifically, ECG amplifies the source concatenation condition early in sampling, then replaces it with edited-frame features while retaining source text and cross-attention conditioning. FAVE complements this guidance with source-sequence and spatial constraints that preserve camera motion and unedited content.}

\rev{During 3D optimization}, existing 3D editing methods~\cite{chen2024gaussianeditor, wang2024view, chen2024dge, wu2024gaussctrl, lee2025editsplat} \rev{can} suffer from global attribute leakage, where gradients intended for the target object erroneously update the entire scene. \rev{Explicit spatial control keeps the refinement localized to the intended target.} To address this issue, we introduce Region-Selective 3DGS Refinement (RS-GS), a method that explicitly constrains the refinement to the target region by back-projecting 2D segmentation masks into 3D space and using depth-guided filtering to select the corresponding Gaussians. By confining the optimization to these back-projected primitives while freezing all others, RS-GS ensures that geometric and appearance updates are confined to the edited regions. 

In summary, \rev{VidEdit3D} effectively reformulates 3D scene editing as a trajectory-based video editing task.
Our main contributions are:
\begin{itemize} 
\item \rev{We propose VidEdit3D, a framework that reformulates 3D editing as video editing to propagate a reference edit consistently across viewpoints.}
\item We design TGR to \rev{select informative keyframes and} organize \rev{them} into smooth \rev{video} clips, improving their compatibility with pretrained I2V priors.
\item We develop FAVE with ECG, a training-free mechanism that \rev{combines scheduled source--edit image conditioning with a separate source-trajectory constraint} to enable precise edit propagation. 
\item We design RS-GS refinement to confine geometric and appearance updates to target primitives, \rev{reducing} background attribute leakage. 
\item \rev{Experiments demonstrate stronger instruction alignment and better multi-view consistency than the compared baselines.}

\end{itemize}

\section{Related Work}
\label{sec:relatedwork}

A significant body of work has explored editing 3D scenes (NeRFs~\cite{mildenhall2021nerf} or 3DGS~\cite{kerbl20233d}) through optimization or distillation strategies.
Early implicit approaches relied on mesh scaffolds or interactive distillation~\cite{yuan2022nerf, jambon2023nerfshop}, while subsequent explicit methods utilize iterative optimization to distill 2D diffusion priors into 3D representations~\cite{chen2024gaussianeditor, koo2024posterior, wang2024gaussianeditor}. 
Recent advancements further enhance quality through identity-preserving noise estimation~\cite{kim2024dreamcatalyst} or task-specific optimization for style transfer and virtual try-on~\cite{liu2024stylegaussian, cao2024gs}. 
However, ensuring multi-view consistency remains a primary bottleneck. 
To address this, several methods \rev{use} geometric \rev{guidance} such as depth or epipolar alignment~\cite{wu2024gaussctrl, chen2024dge, wang2024view, lee2025editsplat, gomel2024diffusion}. 
\rev{Some methods use per-scene fine-tuning~\cite{wang2025personalize,zhuang2024tip}, while others train generalized multi-view editors on paired data~\cite{zhao2025tinker,chen2026omni}.} \rev{ViP3DE~\cite{chen2026fast} propagates an edited view through motion-preserved noise blending and geometry-aware denoising.} \rev{VidEdit3D uses inversion-free source--edit conditioning to drive the requested change, together with spatial source regularization that is weaker inside the edited region and stronger outside it.} \rev{This separates edit transfer from preservation of surrounding content, and our comparisons with ViP3DE show gains in both instruction alignment and multi-view consistency.}

\rev{For image editing,} SDEdit balances realism and fidelity through noising and denoising~\cite{meng2021sdedit}. With \rev{diffusion transformer (DiT)} backbones and rectified flows~\cite{peebles2023scalable,liu2022flow}, \rev{later methods reduce inversion drift through ODE solvers, noise alignment~\cite{deng2024fireflow,wang2024taming,xie2025dnaedit}, or history-guided velocity correction~\cite{wang2026free}. Other methods} preserve structure through attention and feature control~\cite{hertz2022prompt,tumanyan2023plug,liu2024towards}. MasaCtrl supports non-rigid edits with mutual self-attention~\cite{cao2023masactrl}, while KV-Edit and FreeFlux use DiT-specific key-value reuse for background-preserving, task-adaptive editing~\cite{zhu2025kv,wei2025freeflux}. Video editing adds temporal coherence: cross-frame attention control and fusion~\cite{liu2024video,qi2023fatezero} and space-time modulation~\cite{yang2025videograin} reduce drift and improve fine-grained control. I2V methods propagate first-frame edits using temporal features or motion--content alignment~\cite{ku2024anyv2v,liu2025stablev2v}, whereas FlowDirector uses inversion-free ODE steering~\cite{li2025flowdirector}. Complex motion and large appearance or shape changes remain challenging.

\section{Method}
\label{sec:method}

\begin{figure}[t]
\revisionfloat
\centering
\includegraphics[width=\linewidth]{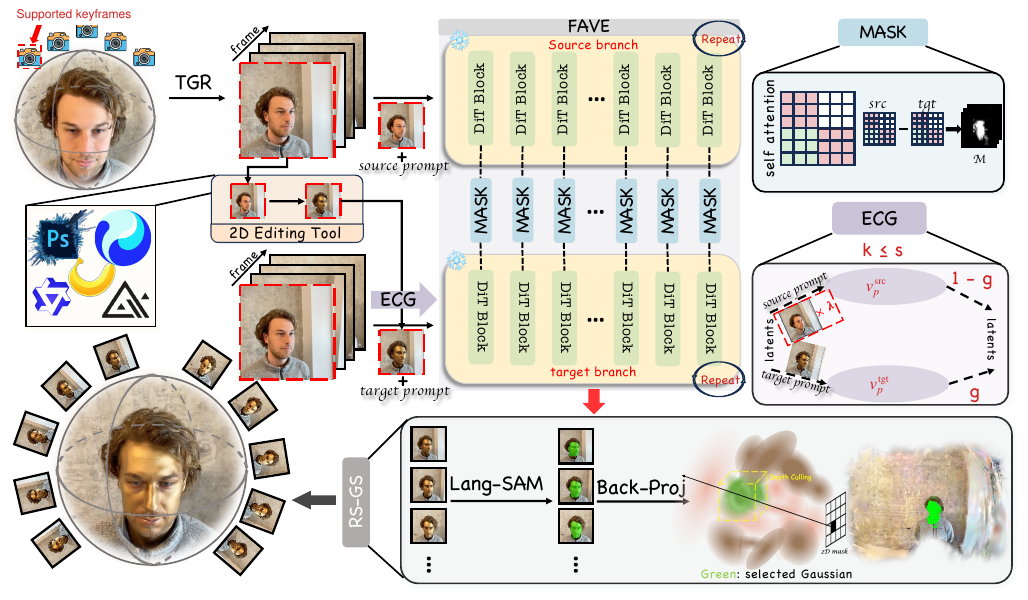}
\caption{Overview of \rev{VidEdit3D}. We reformulate 3D scene editing as a trajectory-based video editing task. The pipeline consists of three key stages: (1) \emph{TGR}: We organize selected views into smooth video clips. (2) \emph{FAVE}: Given a user-edited anchor frame, we propagate the edit across the clip using a pretrained VDM. This module incorporates ECG for semantic alignment and a self-attention-based mask for background preservation. (3) \emph{RS-GS}: We lift the 2D edits back to 3D via inverse rendering to confine updates to the target Gaussians.}
\label{fig:pipeline}
\end{figure}

\rev{Given a reconstructed 3D Gaussian scene, calibrated source views, an edited anchor image, and source and target prompts, VidEdit3D produces an edited 3D scene.} As illustrated in \cref{fig:pipeline}, our pipeline consists of three stages: (1) TGR converts \rev{selected} multi-view renderings into smooth \rev{video} clips; (2) FAVE propagates edits across the clip through a trajectory-regularized flow formulation, augmented with ECG for semantic alignment and a dynamic attention mask for background preservation; and (3) RS-GS lifts the edited frames back into 3D via depth-constrained back-projection, confining updates to the target Gaussians to preserve scene-wide consistency.

\subsection{Preliminaries}
\label{sec:preliminary}

\noindent\textbf{3D Gaussian Splatting.}
3DGS~\cite{kerbl20233d} represents a scene using a set of explicit 3D Gaussians, each parameterized by position $\boldsymbol{\mu}$, covariance $\boldsymbol{\Sigma}$, opacity $\alpha$, and view-dependent color $\mathbf{c}$. Its spatial density is:
\begin{equation}
G(\mathbf{x}) = \exp\!\left(-\frac{1}{2}(\mathbf{x}-\boldsymbol{\mu})^T \boldsymbol{\Sigma}^{-1} (\mathbf{x}-\boldsymbol{\mu})\right).
\end{equation}

Alpha-compositing their projections yields the pixel color $\mathbf{C}$:
\begin{equation}
\rev{\mathbf{C} = \sum_i \mathbf{c}_i \alpha_i \prod_{j=1}^{i-1} (1 - \alpha_j).}
\end{equation}
\rev{Contributing Gaussians are indexed from front to back, and $\alpha_i$ denotes their effective opacity at that pixel.}
Because 3DGS is an explicit primitive-based representation, it is naturally well suited for 3D scene editing.

\vspace{2mm}
\noindent\textbf{Rectified Flow.}
Generative flow models transport a source distribution $p_0$ to a target distribution $p_1$ through the ODE $\mathrm{d}\mathbf{z}_t = \mathbf{v}(\mathbf{z}_t, t)\,\mathrm{d}t$, where $\mathbf{v}$ denotes a neural velocity field~\cite{liu2022flow}. Rectified flows align this transport with the linear interpolation path $\mathbf{z}_t = (1-t)\mathbf{x}_0 + t\mathbf{x}_1$, where $\mathbf{x}_0 \sim p_0$ and $\mathbf{x}_1 \sim p_1$. This trajectory serves as the basis for our flow-based edit propagation in FAVE.
\rev{For FAVE, we use noise time $t\in[0,1]$, with clean data at $t=0$ and noise at $t=1$.}
\subsection{Trajectory-Guided Rendering}
\label{sec:tgr}

To exploit the spatiotemporal priors of video diffusion models for 3D editing, we need to reduce the mismatch between multi-view renderings and the smooth temporal patterns commonly observed in natural videos. \rev{When consecutive camera views are far apart or presented in an irregular order, objects can shift abruptly in the image or reveal different surfaces between frames. An I2V model may then produce inconsistent appearances or shapes for the same object across views.} \rev{TGR addresses this mismatch by selecting informative views and connecting them into smooth camera trajectories for edit propagation.}

We begin with a user-selected anchor view $x_{\mathrm{anc}}$.
\rev{Let $\mathcal V=\{(x_i,T_i)\}_{i=1}^{M}$ denote the calibrated source views, where $x_i$ is an image and $T_i$ its camera, and let $a$ index the anchor. To focus viewpoint coverage on the subject carrying the edit, we select keyframes from a candidate set $\mathcal C$ of relevant source views.}

\rev{We select a compact set of $N$ keyframes by farthest-point sampling (FPS). Starting with $S_1=\{a\}$, each step adds the candidate farthest from its nearest selected view:}
\begin{revision}
\begin{equation}
i_k^\star=\mathop{\arg\max}_{i\in\mathcal C\setminus S_k}\min_{j\in S_k}d(i,j),
\qquad S_{k+1}=S_k\cup\{i_k^\star\}.
\label{eq:tgr_coverage}
\end{equation}
\end{revision}
\rev{The distance $d(i,j)$ combines camera-center distance with the angle between viewing directions, accounting for both camera movement and orientation. Sampling ends when $|S_N|=N$, spreading keyframes across the candidate views while retaining the anchor. The budget $N$ is fixed before sampling, and each keyframe retains its calibrated source pose. Appendix~\ref{app:implementation} specifies the distance and candidate filters.}

\rev{The selected keyframes provide viewpoint coverage, but propagating an edit through a long one-way sequence can accumulate deviations from the anchor. We therefore propagate in both directions from the same edited anchor, reducing the number of sequential transfers needed to reach distant views. To organize these two propagation directions, we connect the keyframes into a closed loop rooted at the anchor. The loop can then be split into two branches that traverse opposite directions and meet at an intermediate keyframe.}

\rev{To connect the selected views with a compact trajectory, we choose the closed ordering with the smallest total camera distance. Let $\Pi(S_N)$ contain cyclic orderings of the selected keyframes. With the anchor fixed at the start, we solve:}
\begin{revision}
\begin{equation}
\pi^\star=\mathop{\arg\min}_{\pi\in\Pi(S_N),\,\pi_0=a}
\sum_{i=0}^{N-1}d(\pi_i,\pi_{i+1}),\qquad \pi_N=\pi_0.
\label{eq:path_cost_main}
\end{equation}
\end{revision}
\rev{The small keyframe set allows exact enumeration of the anchored cyclic orderings. This step preserves the FPS coverage while reducing the total transition cost, including the return to the anchor.}

Given the reordered keyframes, we construct a dense camera trajectory via Kochanek--Bartels spline interpolation~\cite{10.1145/964965.808575}.
\rev{The trajectory is rendered into consecutive clips with shared boundary frames. Both branches begin with the same edited anchor. Within each branch, the edited last frame of one clip conditions the next, maintaining continuity as the edit travels toward the meeting keyframe. Bidirectional propagation thus keeps each transfer chain short while retaining a common reference for the two sides of the trajectory.}

\rev{To keep propagation focused on the source subject, we evaluate the semantic relevance of the rendered source clips before editing. Each branch follows consecutive accepted clips and stops before its first rejected clip, preserving a continuous chain of edited boundary images. Appendix~\ref{app:implementation} provides the filtering and interpolation settings.}

\subsection{Flow-Aligned Video Editing (FAVE)}
\label{sec:fave}

A central challenge in our setting is that the rendered multi-view sequence corresponds to a static 3D scene observed along a camera trajectory. Therefore, during edit propagation, the desired variation should come from viewpoint change alone, while the subject layout, pose, and spatial placement should remain stable across frames. In practice, however, off-the-shelf video editing models\rev{~\cite{yang2025videograin,wang2025videodirector,geyer2023tokenflow,li2025flowdirector,liu2025stablev2v}} often introduce spurious object motion or geometric drift, which is particularly harmful for 3D editing.

To address this issue, we develop Flow-Aligned Video Editing~(FAVE), a training-free propagation framework that adapts trajectory regularization to I2V editing. Our design is inspired by FlowAlign~\cite{kim2025flowalign}, which regularizes editing trajectories toward structurally consistent solutions in image editing. We incorporate its source-consistency principle into a coupled source/target propagation process tailored to video-conditioned denoising, enabling a user-specified edit to propagate along a multi-view video trajectory while suppressing undesired subject motion.

\rev{To preserve the source structure during editing, we couple the source and edited states using the same noise. At sampling step $k$, $\boldsymbol\delta_k=\mathbf p_k-\mathbf q_k$ is the latent-space difference between the edited state $\mathbf p_k$ and noised source state $\mathbf q_k$. It starts at zero and accumulates the edit during sampling. Given the clean source latent $\mathbf x$, shared noise $\boldsymbol\epsilon_k\sim\mathcal N(0,I)$, and decreasing noise level $t_k$, we construct the two states as:}
\begin{revision}
\begin{equation}
\mathbf q_k=(1-t_k)\mathbf x+t_k\boldsymbol\epsilon_k,
\qquad
\mathbf p_k=\mathbf q_k+\boldsymbol\delta_k,
\qquad
\boldsymbol\delta_0=0.
\label{eq:shared_noise}
\end{equation}
\end{revision}
\rev{The backbone velocity prediction $v_q=v_\theta(\mathbf q_k,t_k,c_s,e_s)$ uses source text features $c_s$ and source first-frame image features $e_s$ to provide a reconstruction reference. At the edited state $\mathbf p_k$, ECG combines source and target conditions to obtain the guided prediction $v_p$ (Sec.~\ref{sec:ecg}). We estimate the clean states as $\widehat{\mathbf q}_0=\mathbf q_k-t_kv_q$ and $\widehat{\mathbf p}_0=\mathbf p_k-t_kv_p$, and use their difference to regularize the edit update:}
\begin{revision}
\begin{equation}
\boldsymbol\delta_{k+1}=\boldsymbol\delta_k
+(t_{k+1}-t_k)(v_p-v_q)
+\zeta_k\big(\widehat{\mathbf q}_0-\widehat{\mathbf p}_0\big).
\label{eq:fave_velocity}
\end{equation}
\end{revision}
\rev{Here $\zeta_k$ is a per-step, spatially varying coefficient. The velocity difference drives the requested semantic change, while the clean-state correction limits deviation from the source clip. We retain the user edit at the anchor by restoring the first-frame clean latent after each update.}
Finally, we introduce a dynamic region mask to spatially reweight the
source-consistency regularization in Eq.~\ref{eq:fave_velocity}. \rev{In
video DiTs such as Wan~\cite{wan2025wan}, spatiotemporal patch
tokens are processed within a unified self-attention map.} We observe that
\rev{attention from later-frame tokens concentrates on the first frame}, a
phenomenon we term First-Frame Sink. Based on this observation, we
estimate an edit mask \rev{from attention differences between the source-state branch at $\mathbf q_k$ and the target-conditioned branch at $\mathbf p_k$, using attention to the first and current frames.} We then replace \rev{$\zeta_k$} in Eq.~\ref{eq:fave_velocity} with a
region-adaptive weight:
\begin{revision}
\begin{equation}
\zeta_{k,f}(u)=
\begin{cases}
\zeta_{\mathrm{fg}}, & M_{k,f}(u)=1,\\
\zeta_{\mathrm{bg}}, & M_{k,f}(u)=0
\end{cases},
\label{eq:mask_zeta}
\end{equation}
\end{revision}
where \rev{$M_{k,f}(u)$ indicates whether location $u$ in frame $f$ belongs to the estimated edit region at step $k$, and $\zeta_{\mathrm{fg}}$ and $\zeta_{\mathrm{bg}}$ are the foreground and background coefficients. Appendix~\ref{app:implementation} describes the mask construction.} In this way, the source-consistency constraint
is relaxed where semantic changes are desired, while being strengthened in
layout-preserving regions to preserve the background and 3D geometry.
\subsection{Editing Consistency Guidance}
\label{sec:ecg}
In standard I2V generation, classifier-free guidance (CFG) performs two denoiser evaluations under different text conditions while sharing the same first-frame visual anchor. In our setting, however, this formulation is problematic when computing the guided prediction \rev{$v_p$} for edit propagation. In FAVE, an inversion-free editing method, the two guidance evaluations are intended to play different roles: the target-conditioned prediction should follow the edited first frame and target prompt, whereas the source-conditioned reference should preserve the original scene structure using the source first frame and prompt. Specifically, the first-frame image is injected into the I2V model through two visual conditioning pathways: cross-attention and concatenation. \rev{We denote the first-frame features supplied through these pathways by $\mathbf e^{\mathrm{attn}}$ and $\mathbf e^{\mathrm{cat}}$, respectively.} If both denoiser evaluations are conditioned on the edited first-frame features, the strong first-frame anchoring effect in I2V models causes the two predictions to inherit nearly the same image-driven bias. As a result, the source-to-target discrepancy in the guidance residual \rev{can be} suppressed, \rev{weakening} the \rev{editing} signal\rev{.}
\rev{To obtain $v_p$, ECG contrasts two predictions at the same edited state $\mathbf p_k$. We evaluate $v_p^{\mathrm{tgt}}=v_\theta(\mathbf p_k,t_k,c_t,\mathbf e_{\mathrm{tgt}}^{\mathrm{attn}},\mathbf e_{\mathrm{tgt}}^{\mathrm{cat}})$ with target text features $c_t$ and edited-anchor features, and $v_p^{\mathrm{src}}=v_\theta(\mathbf p_k,t_k,c_s,\mathbf e_{\mathrm{src}}^{\mathrm{attn}},\tilde{\mathbf e}_{\mathrm{src},k}^{\mathrm{cat}})$ as the source-conditioned reference. The source-side reference retains the source text and source cross-attention features throughout sampling. During early denoising ($k\leq s$), we amplify its concatenation condition as $\tilde{\mathbf e}_{\mathrm{src},k}^{\mathrm{cat}}=\lambda\mathbf e_{\mathrm{src}}^{\mathrm{cat}}$; later steps use the edited-frame features $\mathbf e_{\mathrm{tgt}}^{\mathrm{cat}}$. ECG combines these predictions with scale $g$:}
\begin{revision}
\begin{equation}
v_p=v_p^{\mathrm{src}}+g\big(v_p^{\mathrm{tgt}}-v_p^{\mathrm{src}}\big).
\label{eq:ecg}
\end{equation}
\end{revision}

\rev{The conditioning schedule strengthens the source--edit contrast early in denoising while allowing later updates to follow the developing target appearance. The source-conditioned prediction $v_p^{\mathrm{src}}$ is evaluated at $\mathbf p_k$, whereas $v_q$ in \cref{eq:fave_velocity} anchors the update to the source clip through $\mathbf q_k$. Together, these two forms of source conditioning guide reference transfer while regularizing the scene structure.}
\subsection{\rev{Region-Selective 3DGS Refinement}}
\label{sec:rsgs}
\rev{To preserve unedited scene content during 3D fitting, we restrict optimization to the Gaussians that belong to the target region.} We first extract a binary target mask $M_v$ from each view using LangSAM~\cite{ravi2024sam2}. To lift these masks into 3D, we project the center $\boldsymbol{\mu}_i$ of each Gaussian into view $v$, obtaining pixel $\mathbf{u}_{iv}=\pi_v(\boldsymbol{\mu}_i)$ and depth $z_{iv}$. We render the source depth map $D_v$ as a weighted average of Gaussian depths:
\begin{revision}
\begin{equation}
D_v(\mathbf{u})=\frac{\sum_j \rho_{jv}(\mathbf{u})z_{jv}}
{\sum_j \rho_{jv}(\mathbf{u})+\epsilon},
\qquad \rho_{jv}(\mathbf{u})=\alpha_{jv}(\mathbf{u})T_{jv}(\mathbf{u}).
\label{eq:source_depth}
\end{equation}
\end{revision}
where $\rho_{jv}$ is the standard alpha-compositing weight and $T_{jv}$ is the accumulated transmittance\rev{; $\epsilon$ is a small constant for numerical stability}. We retain a projection only if it lies inside the image, has sufficient alpha, and is close to the visible surface, \ie, $|z_{iv}-D_v(\mathbf{u}_{iv})|<\tau_{\mathrm{depth}}$\rev{, where $\tau_{\mathrm{depth}}$ is the depth tolerance}. This depth test prevents background Gaussians behind the target from being selected. Let $\mathcal{V}_i$ denote the valid views of Gaussian $i$. Its target-region support and binary selection state are:
\begin{equation}
w_i =
\frac{1}{|\mathcal{V}_i|}
\sum_{v\in\mathcal{V}_i}M_v(\mathbf{u}_{iv}),
\qquad
m_i=\mathbf{1}[w_i>\tau_w].
\label{eq:gaussian_selection}
\end{equation}

\rev{Here $\tau_w$ is the selection threshold; we set $w_i=0$ if $\mathcal{V}_i$ is empty.} During refinement, $m_i$ gates the gradients of Gaussian position, appearance, scale, rotation, and opacity. \rev{Optimization is restricted to the selected Gaussians, while all remaining parameters stay fixed. New Gaussians inherit their parent's selection state throughout subsequent optimization.}

\section{Experiments}
\label{sec:experiments}

\subsection{Implementation Details}
\label{subsec:protocol}

\rev{We evaluate VidEdit3D on 50 editing cases, comparing against seven baselines and diverse controls.} \rev{We use Wan2.1-I2V-14B-480P~\cite{wan2025wan} with a 50-step schedule at strength 0.9, giving 45 effective denoising steps.} \rev{Settings are scale $g=15$, source-image scale $\lambda=1.5$, condition-switch index $s=30$, and foreground/background regularization coefficients 0.001/0.005.} \rev{The 3D stage uses Nerfstudio Splatfacto~\cite{nerfstudio} with fixed cameras and local Gaussian updates.} \rev{Additional implementation details are provided in Appendix~\ref{app:implementation}.}

We compare \rev{GaussianEditor}~\cite{chen2024gaussianeditor}, DGE~\cite{chen2024dge}, \rev{VCEdit~\cite{wang2024view},} GaussCtrl~\cite{wu2024gaussctrl}\rev{,} EditSplat~\cite{lee2025editsplat}, Omni3DEdit~\cite{chen2026omni}, and ViP3DE~\cite{chen2026fast}. \rev{All methods use the same source scenes and the same editing prompts. Reference-conditioned methods additionally receive the same edited anchor image for each case.}

\rev{CLIP-D measures alignment between image-change and text-change directions, and CLIP-S measures edited-image alignment with the target description~\cite{radford2021learning}.} \rev{DINO measures pairwise DINOv2 feature similarity across edited views~\cite{oquab2023dinov2}.} \rev{MEt3R evaluates geometric feature consistency~\cite{asim25met3r} on a common set of overlapping adjacent camera pairs.}

\subsection{Quantitative Comparisons}

\begin{table}[!htbp]
\revisionfloat
\centering
\small
\setlength{\tabcolsep}{7pt}
\begin{tabular}{lcccc}
\toprule
Method & CLIP-D $\uparrow$ & CLIP-S $\uparrow$ & DINO $\uparrow$ & MEt3R $\downarrow$ \\
\midrule
GaussianEditor & 0.0607 & 0.3001 & 0.8309 & 0.1195 \\
DGE & 0.0747 & 0.3053 & 0.7674 & 0.1259 \\
VCEdit & 0.0214 & 0.3015 & 0.8004 & 0.1203 \\
GaussCtrl & 0.0450 & 0.3023 & 0.8302 & 0.1137 \\
EditSplat & 0.0813 & 0.3061 & 0.7943 & 0.1535 \\
Omni3DEdit & 0.0588 & 0.3010 & 0.7559 & 0.1471 \\
ViP3DE & 0.0134 & 0.3009 & 0.7815 & 0.1220 \\
\rowcolor{oursred} \textbf{VidEdit3D} & \textbf{0.1275} & \textbf{0.3129} & \textbf{0.8327} & \textbf{0.1126} \\
\bottomrule
\end{tabular}
\caption{Quantitative comparison of edit fidelity and multi-view consistency. VidEdit3D achieves the highest CLIP-D, CLIP-S, and DINO and the lowest MEt3R. Bold denotes the best value in each column.}
\label{tab:quantitative_results}
\end{table}

\rev{Table~\ref{tab:quantitative_results} shows that VidEdit3D leads all four metrics, improving edit adherence and multi-view consistency. Because its inference window only contains 10 views~\cite{chen2026omni}, Omni3DEdit has limited cross-view context for maintaining consistent edits along our longer trajectories. VidEdit3D also outperforms inversion-based ViP3DE under the same edited-anchor setting. Appendix~\ref{app:evaluation} provides comparisons using the same Wan backbone. Our user study (Table~\ref{tab:userstudy}) also favors VidEdit3D in multi-view consistency, prompt fidelity, and unedited-region preservation.}

\subsection{Qualitative Comparisons}
\label{subsec:qual_compare}

\begin{figure}[!htbp]
\revisionfloat
\centering

\includegraphics[width=\linewidth]{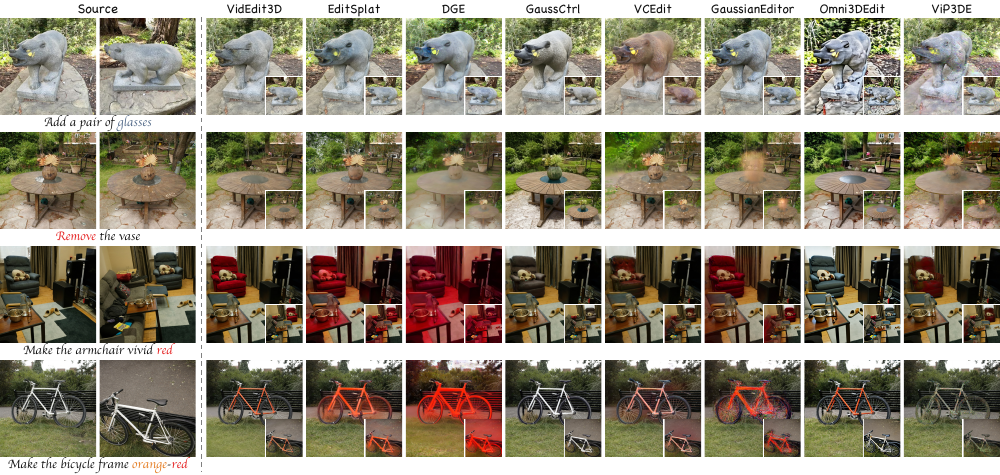}

\caption{Qualitative comparison of reference-guided scene editing. \rev{VidEdit3D}  inserts accessories,{} removes a vase, and localizes appearance changes to the requested objects. Source and edited results are shown at matched viewpoints, with insets providing a second view.}
\label{fig:qualitative_results}
\end{figure}

\rev{Fig.~\ref{fig:qualitative_results} shows  accessory insertion, vase removal, and recoloring. Baselines show incomplete or inconsistent edits and changes to unrelated content; VidEdit3D preserves nearby content and maintains the edits across views. Appendix~\ref{app:more_results} provides additional comparisons and results.}

\subsection{Ablation Study}

\begin{figure}[!htbp]
\revisionfloat
\centering
\includegraphics[width=\linewidth,height=0.84\textheight,keepaspectratio]{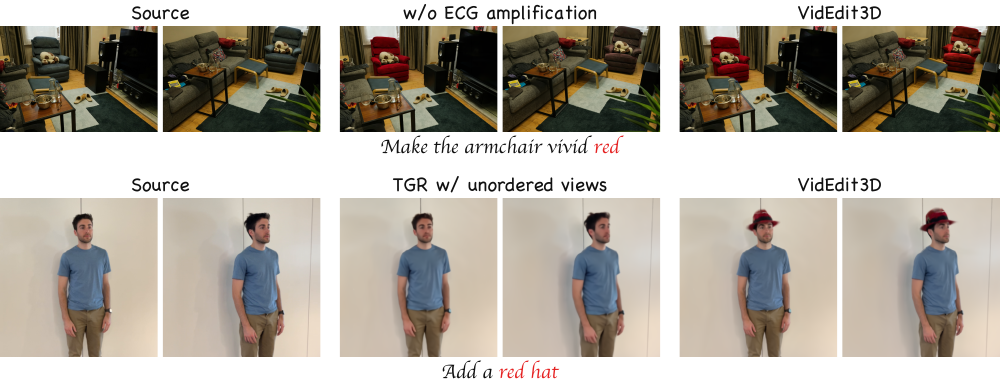}
\caption{Visual effects of ECG amplification and TGR view ordering. ECG amplification is disabled by setting $\lambda=1$, while the unordered TGR variant shuffles the source sequence. \rev{Without amplification, the red-chair edit weakens in the oblique view; VidEdit3D retains the red appearance across both views. Ordered views support hat insertion.}}
\label{fig:ablation}
\end{figure}

\begin{figure}[!htbp]
\revisionfloat
\centering
\includegraphics[width=\linewidth,height=0.84\textheight,keepaspectratio]{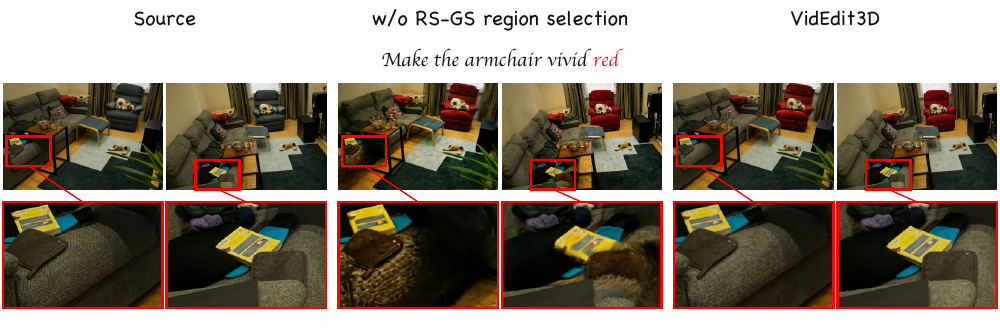}
\caption{Effect of RS-GS region selection. Without restricting Gaussian updates, recoloring the armchair also alters the adjacent sofa. \rev{VidEdit3D} preserves the sofa appearance while completing the red-chair edit. Colored boxes mark corresponding regions enlarged below.}
\label{fig:region_ablation}
\end{figure}

\rev{Table~\ref{tab:detailed_ablation} and Figs.~\ref{fig:ablation} and~\ref{fig:region_ablation} show that source-guided propagation, coherent view organization, and localized 3D refinement together support reliable edits. Replacing source-guided propagation with vanilla I2V or disrupting view order causes the largest losses in edit alignment. VidEdit3D achieves the best scores across all four metrics among the tested variants. The visual results show more complete edits and better preservation of nearby content.}

\begin{table}[!htbp]
\revisionfloat
\centering
\small
\setlength{\tabcolsep}{7pt}
\begin{tabular}{lcccc}
\toprule
Method & CLIP-D $\uparrow$ & CLIP-S $\uparrow$ & DINO $\uparrow$ & MEt3R $\downarrow$ \\
\midrule
w/o ECG amplification & 0.1064 & 0.3116 & 0.8213 & 0.1164 \\
w/o FAVE (vanilla Wan I2V) & -0.0022 & 0.2480 & 0.6693 & 0.1838 \\
w/o RS-GS region selection & 0.1246 & 0.3116 & 0.8276 & 0.1148 \\
w/o RS-GS depth culling & 0.1260 & 0.3116 & 0.8309 & 0.1137 \\
w/o FAVE source regularization & 0.1270 & 0.3116 & 0.8149 & 0.1187 \\
TGR w/ unordered views & 0.0405 & 0.3001 & 0.8085 & 0.1200 \\
TGR w/ pose-sorted views & 0.0631 & 0.3043 & 0.8042 & 0.1199 \\
TGR w/ LERP + SLERP & 0.1242 & 0.3116 & 0.8309 & 0.1143 \\
\rowcolor{oursred} \textbf{VidEdit3D} & \textbf{0.1275} & \textbf{0.3129} & \textbf{0.8327} & \textbf{0.1126} \\
\bottomrule
\end{tabular}
\caption{Ablation of FAVE, ECG, RS-GS, and TGR. VidEdit3D achieves the best scores across all four metrics. Bold denotes the best value in each column.}
\label{tab:detailed_ablation}
\end{table}

\begin{samepage}
\section{Conclusion}
\label{sec:conclusion}
We presented \rev{VidEdit3D}, which reformulates \rev{reference-guided} 3D scene editing as I2V propagation by bridging static multi-view observations with the spatiotemporal priors of pretrained video diffusion models. TGR organizes \rev{selected} 3D views into coherent video-like streams \rev{for edit propagation}. FAVE enables robust semantic propagation through an inversion-free flow formulation, while ECG \rev{uses scheduled image conditioning to maintain} a clearer source-to-target direction \rev{in} DiT-based I2V models. RS-GS confines updates to target regions during final 3D optimization, reducing background leakage and better preserving unedited scene content. Together, these components enable high-fidelity, view-consistent edits across diverse viewpoints and challenging structural modifications without \rev{network} fine-tuning. \rev{VidEdit3D shows how pretrained video priors support flexible 3D editing.}
\par
\end{samepage}

\FloatBarrier
{\small
\let\originalbibitem\bibitem
\let\bibitem\revisionbibitem
\bibliographystyle{ieeenat_fullname}
\bibliography{references}}
\clearpage
\appendix
\begin{center}{\Large\bfseries Supplementary Material}\end{center}
\begingroup
\setlength{\abovedisplayskip}{4pt plus 2pt minus 1pt}
\setlength{\belowdisplayskip}{4pt plus 2pt minus 1pt}
\setlength{\abovedisplayshortskip}{0pt plus 2pt}
\setlength{\belowdisplayshortskip}{4pt plus 2pt minus 2pt}
\section{\rev{Additional Implementation Details}}
\label{app:implementation}

\rev{The edited anchor conditions the I2V backbone, while the rendered source clip guides scene preservation. Source and edited states share noise at each denoising step; ECG amplifies source concatenation features early in sampling. Clips contain 61 frames to match the causal autoencoder's temporal compression and retain their rendering camera intrinsics and field of view during propagation.}

\rev{We use $N=4$ keyframes for small scenes and $N=7$ for large scenes, including the user-selected anchor in both cases. The keyframe count is fixed for each scene. Candidate views are screened using CLIP ViT-H/14 similarity to a description of the source subject carrying the edit. A view is retained when its similarity is at least 0.15 and at least 0.74 times the source anchor similarity. For example, adding glasses uses a description of the person's head. We also project the reconstructed subject points into each candidate camera and require at least 90\% of these points to lie in front of the camera and inside the image. We exclude views where the projected subject occupies too little or too much of the image, or lies too far from the image center. Duplicate camera poses are removed.}

\rev{For FPS (Eq.~\ref{eq:tgr_coverage}) and closed-loop ordering (Eq.~\ref{eq:path_cost_main}), we use $d(i,j)=\|\mathbf c_i-\mathbf c_j\|_2+\arccos(\mathbf f_i^\top\mathbf f_j)$, where $\mathbf c_i$ is the camera center in the scene coordinates used for reconstruction and $\mathbf f_i$ is the unit viewing direction. The angular term is measured in radians, and both terms have unit weight. Kochanek--Bartels interpolation uses tension 0.5, with 61 samples per segment and a shared endpoint between consecutive segments.}

\rev{Before editing, we score every unedited rendered frame with the same source-subject description. A clip is accepted when its minimum frame similarity is at least 0.74 times that of the original anchor render; the anchor score must also reach 0.15. Both propagation branches use this same anchor reference, and each stops at its first rejected clip. Within each accepted branch, the edited last frame of one clip conditions the next.}

\rev{The edit mask $M_{k,f}(u)$ in Eq.~\ref{eq:mask_zeta} compares attention from the source-state branch and the target-conditioned edited-state branch, restricted to the first and current frames. We average the normalized differences across the 40 DiT blocks and threshold at 0.25. The mask is recomputed at each denoising step and upsampled to assign the foreground and background regularization weights. The edited anchor is restored directly.}

\rev{Starting from the source 3DGS, we optimize the selected Gaussians for 3,000 iterations using the edited frames and fixed cameras. The objective combines an $\ell_1$ term with structural similarity (SSIM)~\cite{wang2004ssim}:}
\begin{revision}
\begin{equation}
\mathcal L_{\mathrm{3DGS}}=0.8\|\widehat I_v-\widetilde I_v\|_1
+0.2\big(1-\mathrm{SSIM}(\widehat I_v,\widetilde I_v)\big),
\end{equation}
\end{revision}
\rev{where $\widehat I_v$ and $\widetilde I_v$ are the rendered and edited target images. Unselected Gaussian parameters remain fixed.}

\section{\rev{Additional Evaluation}}
\label{app:evaluation}

\rev{CLIP-D and CLIP-S use CLIP ViT-B/32~\cite{radford2021learning}.} Let \rev{$\hat i_b,\hat i_a$ be normalized image features} before and after editing, and \rev{$\hat t_s,\hat t_t$ normalized} source and target text \rev{features}. \rev{We compute:}
\begin{revision}
\begin{equation}
\mathrm{CLIP\!\!\!-D}=\cos(\hat i_a-\hat i_b,\hat t_t-\hat t_s),
\qquad
\mathrm{CLIP\!\!\!-S}=\cos(\hat i_a,\hat t_t).
\end{equation}
\end{revision}
\rev{CLIP-D averages views with nonzero change directions. For $n$ edited views, let $d_i$ denote the normalized DINOv2 ViT-B/14 CLS feature of view $i$~\cite{oquab2023dinov2}. We average their pairwise cosine similarity:}
\begin{revision}
\begin{equation}
\mathrm{DINO}=\frac{1}{n(n-1)}\sum_{i\ne j}\cos(d_i,d_j).
\end{equation}
\end{revision}
\rev{MEt3R~\cite{asim25met3r} compares image features after geometric reprojection on the same overlapping adjacent camera pairs for all methods; lower values indicate better consistency. We average each metric within a case, then over cases.}

\begingroup
\setlength{\intextsep}{6pt}
\begin{table}[H]
\revisionfloat
\centering
\small
\setlength{\tabcolsep}{7pt}
\begin{tabular}{lcccc}
\toprule
Method & CLIP-D $\uparrow$ & CLIP-S $\uparrow$ & DINO $\uparrow$ & MEt3R $\downarrow$ \\
\midrule
ViP3DE-Wan & 0.0472 & 0.3017 & 0.7946 & 0.1193 \\
FlowAlign-Wan & 0.0970 & 0.3083 & 0.8079 & 0.1186 \\
\rowcolor{oursred} \textbf{VidEdit3D} & \textbf{0.1275} & \textbf{0.3129} & \textbf{0.8327} & \textbf{0.1126} \\
\bottomrule
\end{tabular}
\caption{Comparison using the same Wan2.1-I2V-14B-480P backbone. Bold denotes the best value in each column.}
\label{tab:wan_comparison}
\end{table}
\endgroup

\rev{Table~\ref{tab:wan_comparison} compares VidEdit3D with Wan adaptations of ViP3DE and FlowAlign~\cite{kim2025flowalign}. \emph{ViP3DE-Wan} replaces SVD while retaining ViP3DE's inversion-based initialization, motion-preserved noise blending, 3D correspondences, and conditional-branch feature override. ViP3DE injects first-frame features into 3D-corresponding positions at the final SVD U-Net layer during every denoising step~\cite{chen2026fast}; we transfer this intervention to its direct DiT counterpart. The edited anchor, camera views, resolution, sampling steps, and random seeds remain identical. FlowAlign-Wan adapts FlowAlign to Wan2.1-I2V-14B and uses the same edited anchor, TGR trajectories, and RS-GS refinement as VidEdit3D.}

\endgroup
\FloatBarrier

\section{\rev{Additional Qualitative Results}}
\label{app:more_results}

\subsection{\rev{Additional Comparisons}}

\rev{Fig.~\ref{fig:additional_comparisons} compares lettering, accessory insertion, and local recoloring across  three{} scenes. VidEdit3D keeps the  headrest lettering legible, maintains the red bandana across views, and confines the white recoloring to the plant leaves.}

\begingroup
\setlength{\intextsep}{6pt}
\begin{figure}[!htbp]
\revisionfloat
\centering
\captionsetup{skip=3pt}

\includegraphics[width=\linewidth]{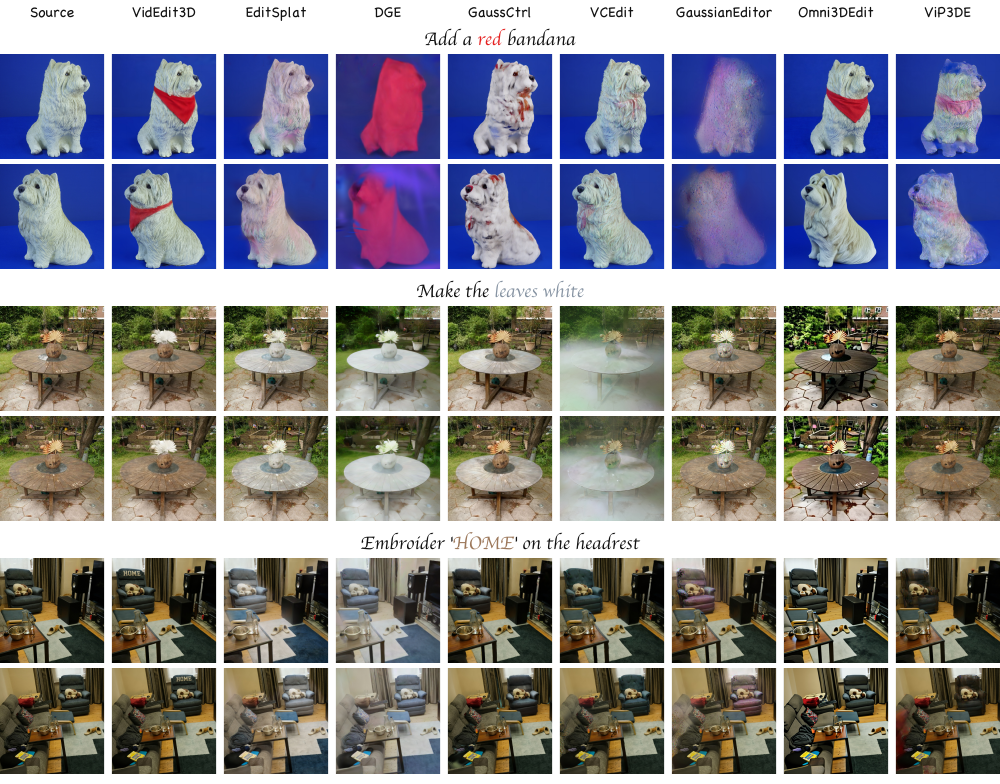}

\caption{Additional qualitative comparisons of text transfer, accessory insertion, and localized recoloring. Each case shows two views of VidEdit3D and the compared baselines.}
\label{fig:additional_comparisons}
\end{figure}
\endgroup

\FloatBarrier

\subsection{\rev{More Results}}

\begin{figure}[!htbp]
\revisionfloat
\centering
\captionsetup{skip=3pt}
\includegraphics[width=\linewidth]{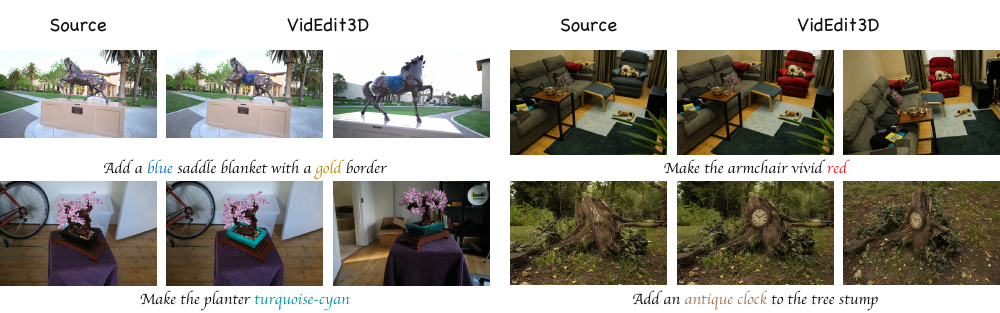}
\par\nointerlineskip
\includegraphics[width=\linewidth]{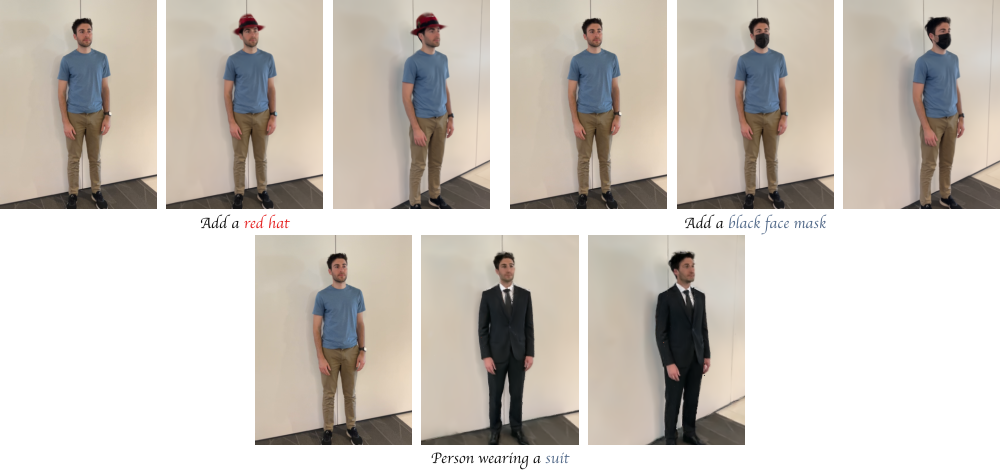}
\caption{More results. Each task shows a source image followed by two edited views.}
\label{fig:more_results_objects}
\end{figure}

\begin{figure}[!htbp]
\revisionfloat
\centering
\captionsetup{skip=3pt}
\includegraphics[width=\linewidth]{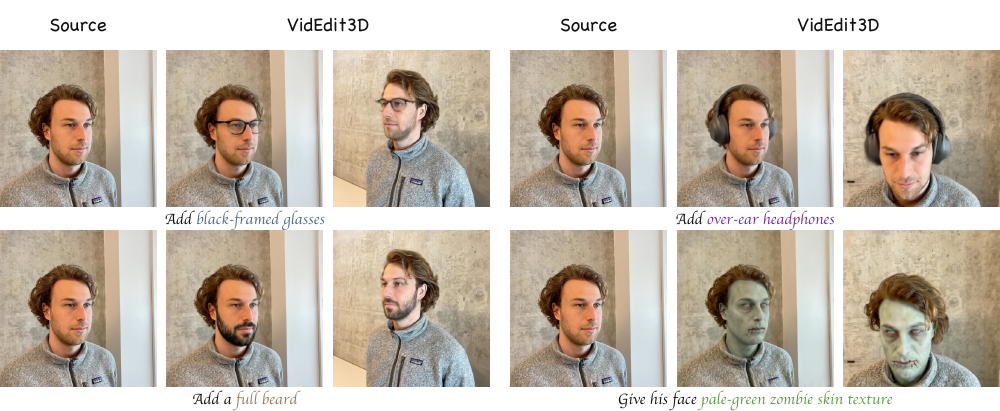}
\par\nointerlineskip
\includegraphics[width=\linewidth]{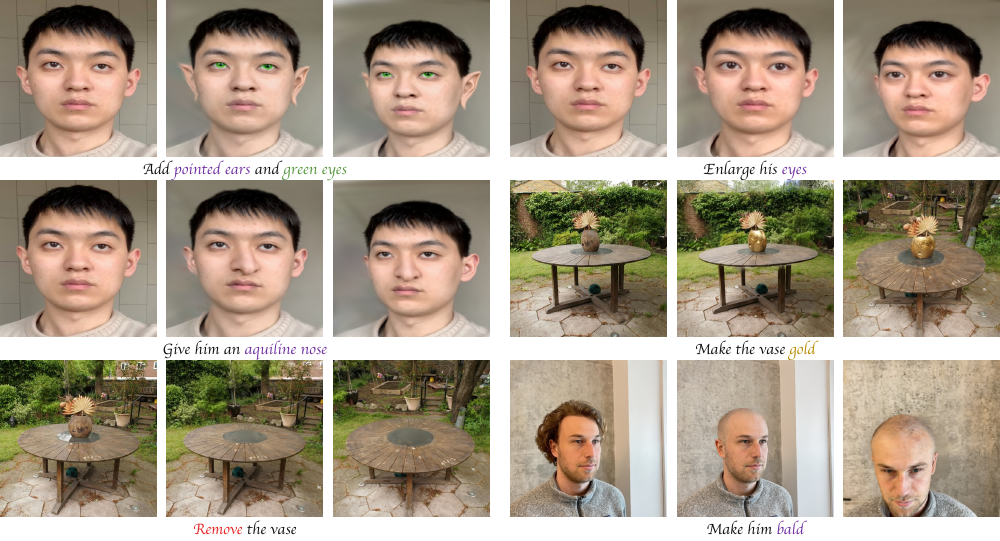}
\caption{More results (continued). Each task shows a source image followed by two edited views.}
\label{fig:more_results_appearance}
\end{figure}

\begin{figure}[H]
\revisionfloat
\centering
\captionsetup{skip=3pt}
\includegraphics[width=\linewidth]{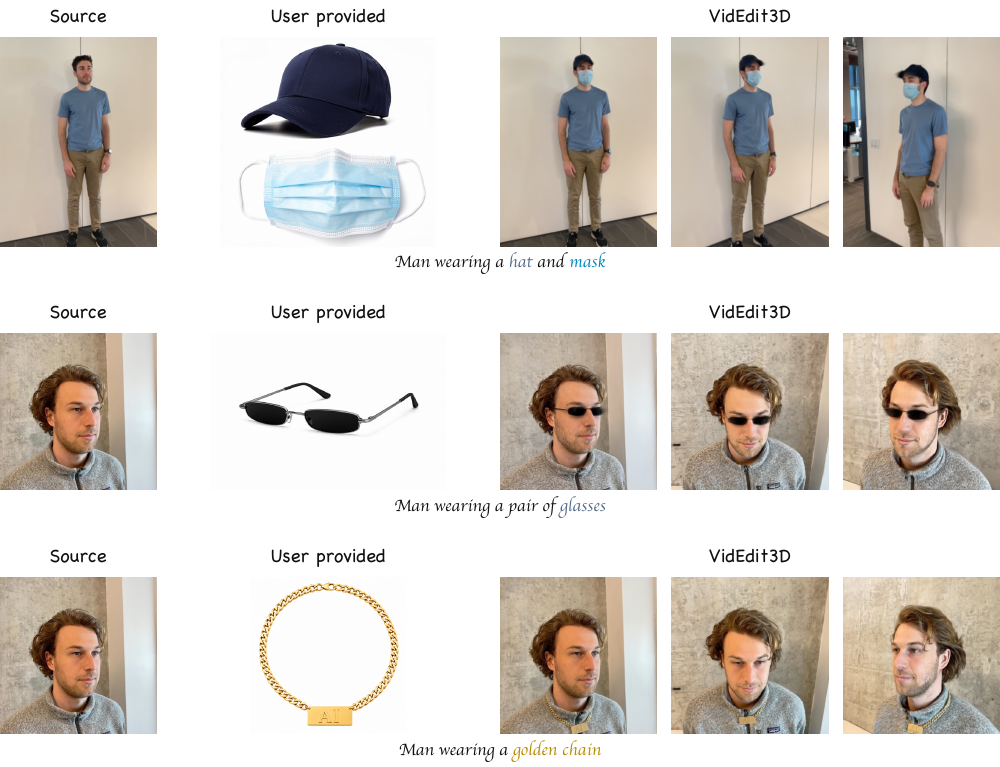}
\caption{Edits with user-provided accessory references. User-provided images specify the hat and mask, sunglasses, and AI pendant used in the edited anchor. Each row shows the source, the reference image, and three views rendered from the edited 3D scene.}
\label{fig:reference_guided_results}
\end{figure}

\rev{Figs.~\ref{fig:more_results_objects} and~\ref{fig:more_results_appearance} demonstrate object insertion and removal, local recoloring, and changes to clothing and facial attributes across multiple viewpoints. VidEdit3D also accepts user-provided accessory images to specify the desired appearance (Fig.~\ref{fig:reference_guided_results}). These references guide the 2D anchor edit, which is then propagated across views and lifted into the 3D scene.}

\FloatBarrier

\section{\rev{User Study}}
\label{app:userstudy}

\rev{Results from each method were presented in randomized order with method names hidden. Participants rated multi-view consistency, prompt fidelity, and unedited-region preservation on a three-point scale (1/2/3, higher is better). VidEdit3D achieves the highest mean score on all three criteria in Table~\ref{tab:userstudy}.}

\begingroup
\setlength{\intextsep}{6pt}
\begin{table}[H]
\revisionfloat
\centering
\small
\setlength{\tabcolsep}{7pt}
\begin{tabular}{lccc}
\toprule
Method & Multi-view consistency & Prompt fidelity & Unedited-region preservation \\
\midrule
GaussianEditor & $1.98\pm0.83$ & $1.63\pm0.76$ & $1.84\pm0.80$ \\
DGE & $2.05\pm0.83$ & $1.68\pm0.85$ & $1.83\pm0.81$ \\
VCEdit & $1.95\pm0.86$ & $1.70\pm0.83$ & $1.80\pm0.78$ \\
GaussCtrl & $1.76\pm0.83$ & $1.60\pm0.78$ & $1.82\pm0.81$ \\
EditSplat & $2.04\pm0.83$ & $1.60\pm0.74$ & $1.95\pm0.76$ \\
Omni3DEdit & $1.74\pm0.82$ & $1.38\pm0.76$ & $1.75\pm0.79$ \\
ViP3DE & $1.52\pm0.81$ & $1.51\pm0.73$ & $1.53\pm0.72$ \\
\rowcolor{oursred} \textbf{VidEdit3D} & $\mathbf{2.53\pm0.72}$ & $\mathbf{2.60\pm0.70}$ & $\mathbf{2.52\pm0.74}$ \\
\bottomrule
\end{tabular}
\caption{User ratings (mean $\pm$ standard deviation). Bold denotes the highest mean score in each column.}
\label{tab:userstudy}
\end{table}
\endgroup

\section{Limitations and Failure Cases}
\label{app:limitations}

\subsection{Failure-Mode and Scope Analysis}
\begin{figure}[H]
    \revisionfloat
\centering
    \includegraphics[width=0.95\linewidth]{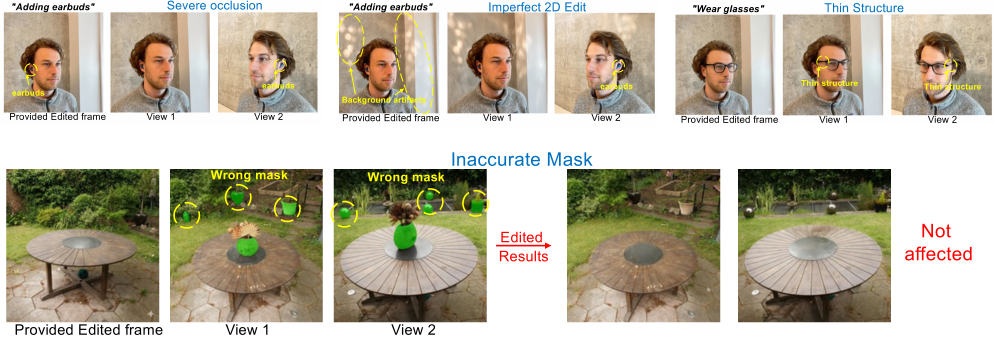}
    \caption{Failure-mode and scope analysis: occlusion, imperfect anchors, thin structures, and segmentation.}
    \label{fig:failure_modes}
\end{figure}

Fig.~\ref{fig:failure_modes} analyzes \rev{inaccurate} segmentation masks, thin structures, severe occlusion, and imperfect 2D anchor edits. \rev{Severe occlusion and ambiguous prompts can make anchor selection and semantic propagation less reliable, whereas thin structures and inaccurate masks stress precise localization. The removal example also shows that extra mask regions need not cause visible changes outside the intended edit.}

\rev{The edited anchor determines the visual content transferred to the 3D scene. Background artifacts and incomplete object removal in the anchor can therefore persist after propagation and 3D refinement. A clean, target-visible anchor reduces these errors, and stronger 2D editors can be incorporated without retraining the propagation model. RS-GS preserves the attributes of unselected Gaussians, although their rendered visibility can change as the edited target changes shape.}

\subsection{Large-Area Appearance Edits}
\begin{figure}[!htbp]
\revisionfloat
\centering
\includegraphics[width=\linewidth,height=0.88\textheight,keepaspectratio]{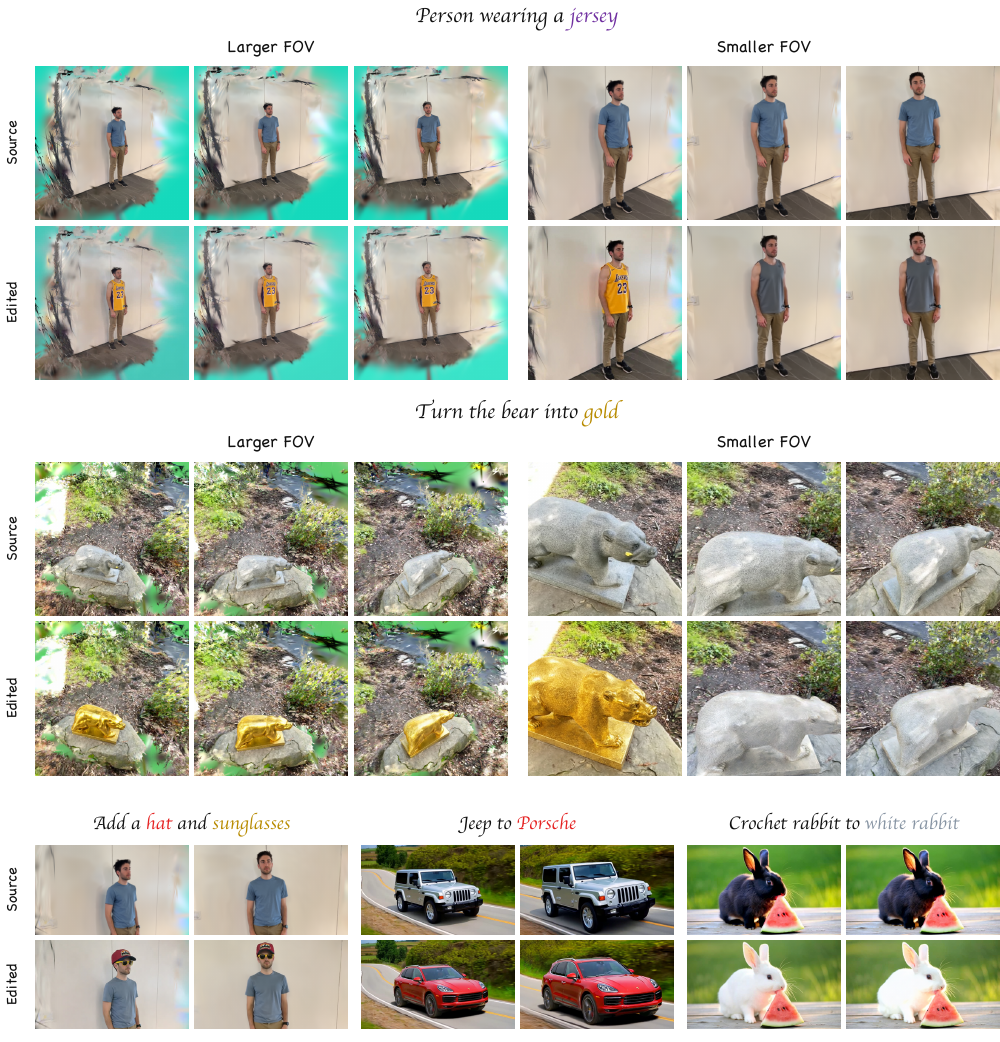}
\caption{Field of view affects the balance between unchanged context and target resolution. Wider views retain surrounding context and better sustain the jersey and gold-material edits, at the cost of target resolution. The localized 3D edit (bottom left) remains stable. Video editing (bottom middle and right) is unaffected by this appearance fading: temporal continuity provides stronger frame-to-frame correspondence and better matches the video diffusion model's learned prior, allowing the target appearance to persist. Rows show source and edited frames.}
\label{fig:tgr-4}
\end{figure}

\rev{Large-area texture or style edits leave little unchanged context to anchor propagation. FAVE uses source-consistency regularization to preserve scene layout, but transferring a new appearance across substantial visibility changes becomes harder when most visible surfaces are edited. The target appearance can then drift or fade toward the source appearance in later views. Fig.~\ref{fig:tgr-4} illustrates this dependence on scene context.}

\rev{A wider rendering field of view introduces more unedited context and can help sustain the target appearance along the trajectory. This reduces the target's image resolution, however, making fine-grained appearance transfer less accurate after 3D refinement. Balancing contextual support with target resolution remains a direction for improving large-area editing.}

\subsection{Editing Time}
\rev{Computational efficiency remains a limitation. Our Wan2.1-I2V-14B backbone requires repeated denoiser evaluations over dense multi-view clips, making interactive editing costly. Sparse attention~\cite{xi2025sparse, yang2025sparse, liu2025fpsattention}, feature caching~\cite{liu2025timestep, zhou2025less}, model distillation~\cite{yin2024improved}, and learned few-step solvers~\cite{zhao2026dyweight} offer promising ways to reduce this cost. Integrating these techniques while preserving reference fidelity and multi-view consistency is an important direction for future work.}

\end{document}